# Natural Language Processing using Hadoop and KOSHIK

**Hong Shi, Emre Erturk**

Eastern Institute of Technology, New Zealand

## Abstract

Natural language processing, as a data analytics related technology, is used widely in many research areas such as artificial intelligence, human language processing, and translation. At present, due to explosive growth of data, there are many challenges for natural language processing. Hadoop is one of the platforms that can process the large amount of data required for natural language processing. KOSHIK is one of the natural language processing architectures, and utilizes Hadoop and contains language processing components such as Stanford CoreNLP and OpenNLP. This study describes how to build a KOSHIK platform with the relevant tools, and provides the steps to analyze wiki data. Finally, it evaluates and discusses the advantages and disadvantages of the KOSHIK architecture, and gives recommendations on improving the processing performance.

## 1. Introduction

### 1.1. Natural Language Processing

Natural language processing (NLP) is a technique to analyze readable text that is generated by humans for artificial intelligence, language processing, and translation (Behzadi, 2015). In order to accurately analyze the text, there are some methods for NLP to use in order to deal with the challenges such as the collection and storage of the text corpus, and analysis. NLP techniques also gain experience and benefits through research in linguistics, computational statics, artificial intelligence, machine learning and other sciences (Behzadi, 2015). However, at present, because of the information explosion, the use of traditional NLP faces many challenges such as the volume of structured and unstructured data, velocity of processing data, accuracy of the results. In addition, there are many slangs and ambiguous expressions used on social media networks, which give NLP pressure to analyze the meanings, which may also be hard for some people. Moreover, people nowadays heavily depend on search engines like Google and Bing (which use NLP as their core technique) in their daily study, work, and entertainment. All of these factors encourage computer scientists and researchers to find more robust, efficient and standardized solutions for NLP.

### 1.2. Big Data

Big Data is designed as generic platform to resolve the issues of volume, velocity, variety, veracity and value in data analytics (IBM, 2012). The data is collected from different sources, for example, daily logs, social media, and business transactions. Big Data demands the ability to hoard large amounts of data. With advanced storage technologies, the data size could be as high as terabytes ($10^{12}$ bytes), petabytes ($10^{15}$ bytes) and exabytes ($10^{18}$ bytes). Furthermore, research on NLP often overlaps with research on or the use of Big Data platforms. Big Data handles all types of formats which are structured and unstructured. The structured data is readable and well-designed that is commonly stored in traditional relationship databases. Unstructured data does not have a predefined format. This type of data can be found in, for example, emails, images, and videos. Big Data is widely used in business forecasts, scientific research, analysis of social issues, healthcare, and meteorology. While promoting advances in NLP, it is also important to be aware of ethical issues around the potential misuse and dual-use of big data and NLP tools (Hovy & Spruit, 2016).

Some of these issues can be very interesting for information technology students to debate and learn more about (Erturk, 2013).

## 1.3. Hadoop

Hadoop is a application created using Java, and provides a set of tools to do data processing which includes data storage, access, analysis (White, 2012). According to White (2012), the main components in Hadoop are HDFS (Hadoop distributed file system) and MapReduce. HDFS is solid and fault-tolerant, and provides a Java-based API that integrates with MapReduce to process the large data in parallel using a cluster of servers (Taylor, 2010). As Murthy, Padmakar and Reddy (2015) observed, traditional relational databases, which store structured data, could also be used together with Hadoop HDFS with full database management systems (DBMS) features. MapReduce is created with inspiration from the theory published by Google, and it provides functions that split large set of data computing into small computing tasks (White, 2012).

## 1.4. Natural Language Processing on Hadoop

Idealy, Hadoop, with the features of distributed storage system, multi-tasks processing system, generic platform and open source, can be used for NLP research. There are many papers discussing the solutions which utilize Hadoop for NLP. For example, HDFS is used in the research of Markham, Kowolenko and Michaelis (2015) to manage large amount of unstructured data, which was collected from the Internet with Hadoop tools. Idris, Hussain, Siddiqi, Hassan, Bilal and Lee (2015) designed a system named MRPack which gives an end-to-end MapReduce processing model for text processing, which includes a job task for NLP. MRPack shows it has better performance in data accessing, data managing and data writing, as well as less programming work and demanding for I/O management (Idris et al., 2015).

This study reviews KOSHIK, a Hadoop based framework which has been developed from the paper written by Exner and Nugues (2014). Then it gives the findings for what Hadoop components this framework used, and recommendations to improve the NLP performance. Finally, it shows the steps to test KOSHIK.

## 2. Literature Review

## 2.1. Introduction

It is described that "KOSHIK is a framework for batch-oriented large scale-processing and querying of unstructured natural language documents. In particular, it builds on the Hadoop ecosystem and takes full advantage of the data formats and tools present in this environment to achieve its task" (Exner & Nugues, 2014, p. 463). KOSHIK tries to resolve the common challenges in NLP such as volume, velocity, and variety by adopting Hadoop for the system infrastructure, using a batch-oriented annotation model to continually add annotations and enabling a generic algorithm platform to analyze the variety of text (Exner & Nugues, 2014). It is also argued that before developing KOSHIK, there were many other NLP frameworks such as MULTEXT, GATE, and UIMA, which were important for document retrieval, indexing, and querying of processed information applications (Exner & Nugues, 2014).

## 2.2. KOSHIK Architecture

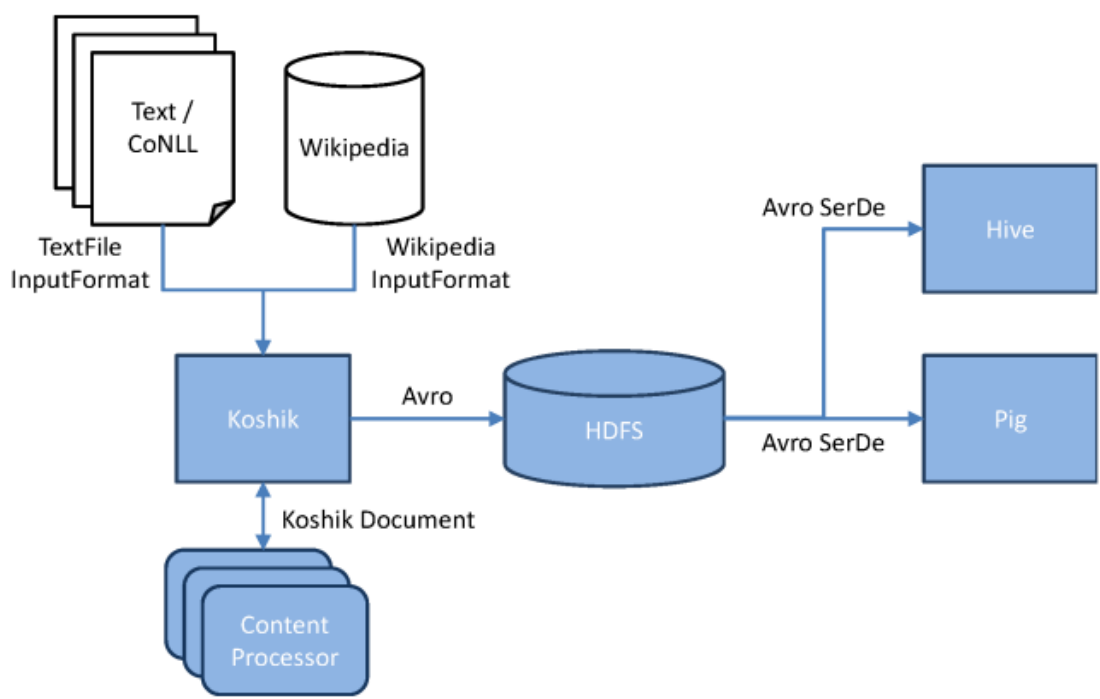

Figure 1. An overview of the KOSHIK architecture. (Exner & Nugues, 2014).

Exner and Nugues (2014) stated that "KOSHIK supports a variety of NLP tools implemented atop of the Hadoop distributed computing framework. The requirements on KOSHIK were driven by the desire to support scalability, reliability, and a large number of input formats and processing tasks". KOSHIK supports different kinds of documents such as CoNLL-X, CoNLL 2008 and 2009 and Wikipedia dump files (Exner & Nugues, 2014; Buchholz and Marsi, 2006; Surdeanu et al., 2008). In order to analyze different kinds of language, KOSHIK involves a large set of filters, tokenizers, taggers, parsers, and coreference solvers through several language specific NLP tools such as OpenNLP, Mate Tools, Stanford CoreNLP, etc. (Exner & Nugues, 2014). With the developed annotation model in KOSHIK, this framework could simplify the process of content processors, which only require computing resource focus on the input and output of annotated documents (Exner & Nugues, 2014).

KOSHIK utilizes Hadoop as its distributed computing framework, which not only used for a better performance on large corpus analysis, but also for the division and distribution of a set of documents, management of processing jobs, and to get the process result (Exner & Nugues, 2014). Pig and Hive which are core members of Hadoop ecosystem are adopted into this framework to manage the workflows for processing large data-sets and easily query information by SQL-like language respectively (Exner & Nugues, 2014).

## 3. Key Terms

After evaluating this paper, Hive and Pig which are data querying components of Hadoop, and OpenNLP and Stanford CoreNLP which are NLP process tools are identified for future research.

According to Thusoo et al. (2009), Hive is a data warehouse infrastructure built based on Hadoop for providing data summarization, query, and analysis. It could be used to deal with large distributed data such as HDFS and traditional file systems like FAT, NTFS and exFat (Thusoo et al., 2009). In addition, HiveQL, a SQL-like language, is created to query data on Hadoop which will translate the query to MapReduce jobs for high performance (Thusoo et al., 2009).

The Pig is a high-level declarative querying language inspired by SQL which will translate the query to MapReduce jobs (Pis Olston, Reed, Srivastava, Kumar & Tomkins, 2008). According to Gates et al. (2009), there are many data analysis projects which adopted Pig because it is easy to learn and use, and could quickly implement different versions of algorithms.

Lingad, Karimi and Yin (2013) found that "OpenNLP is a Java based library for various natural language processing tasks, such as tokenization, part-of-speech (POS) tagging, and named entity recognition. For named entity recognition, it trains a Maximum Entropy model using the information from the whole document to recognize entities in documents". Although OpenNLP provides many functions for NLP, the model to process the document should be considered that Verspoor (2012) argued that OpenNLP has low quality to divide sentence into parts, but it could improve the performance by using annotator.

Stanford CoreNLP is a NLP tool and utilizes annotation to analyse text that it provides most of the common core NLP steps, from tokenization through to coreference resolution (Manning, 2014). It is also described that Stanford CoreNLP provides a complete toolkit and tools for grammatical analysis, accurate analysis, and supports different kinds of languages (Manning, 2014).

## 4. Testing KOSHIK

### 4.1. Preparing the Hardware Environment

According to Exner and Nugues (2014), KOSHIK runs on Hadoop with Hive, and requires other NLP process libraries (OpenNLP, Mate and Stanford CoreNLP). The experimental environment (including hardware, software, and test data) used in this paper is described in the following tables.

| Hardware Environment | |
|---|---|
| Hardware | Description |
| CPU | INTEL i5 quad cores 2.7Ghz |
| Hard disk | SATA 500G |
| Memory | DDR3 16G |
| Graphic Card | NVidia 940 |
| Network | 1G LAN |

| Software Environment | |
|---|---|
| Operation System/application/library | Version |
| Operation System | Centos 6.7 |
| KOSHIK | 1.01 |
| JAVA | 1.80 |
| Hadoop | 2.60 |
| Hive | 1.1.0 |
| Cloudera | 5.70 |
| Hue | 3.90 |
| VirtualBox | 5.0.20 |
| Stanford CoreNLP | 3.6 |
| OpenNLP | 1.5.0 |
| Mate | |

| Test Data | |
|---|---|
| Item | Description |
| Wiki Data | 12GB English wiki data. |

In this test environment, Cloudera Quickstart VM is utilized to quickly construct the software environment. According to Cloudera (2016), Cloudera Quickstart VM is a virtual machine which installed most of the applications related to Hadoop such as Hive, Spark, HBase. One benefit to use this virtual machine is that it has already configured Hadoop and integrated with Hive, HBase and Hue, which is a management tool for applications in Hadoop ecosystem (Cloudera, 2016). Therefore, it is an ideal platform for researching and making proof of concept.

## 4.2. Preparing the Software

### 4.2.1. Download software and wiki data

The following table list the download links for the required software.

| Software | Link |
| --- | --- |
| KOSHIK | https://github.com/peterexner/KOSHIK |
| Cloudera Quickstart VM | http://www.cloudera.com/downloads/quickstart_vms/5-7.html |
| VirtualBox | https://www.virtualbox.org/wiki/Downloads |
| Stanford CoreNLP | http://nlp.stanford.edu/software/corenlp.shtml |
| OpenNLP | http://opennlp.sourceforge.net/models-1.5/ |
| Mate | https://code.google.com/p/mate-tools/downloads/list |
| WikiData | https://dumps.wikimedia.org/enwiki/20160501/enwiki-20160501-pages-articles.xml.bz2 |

### 4.2.2. Configure software

The following table describes steps to configure the required software.

| Software | Step Description |
| --- | --- |
| VirtualBox | 1. Unzip Cloudera Quickstart VM.<br>2. Import the unzipped Cloudera Quickstart VM into VirtualBox. |
| KOSHIK required libraries | 1. Create a folder anywhere named model.<br>2. Create a folder named is2 in model folder and put downloaded files CoNLL2009-ST-English-ALL.anna-3.3.lemmatizer.model, CoNLL2009-ST-English-ALL.anna-3.3.parser.model, CoNLL2009-ST-English-ALL.anna-3.3.postagger.model in it.<br>3. Create a folder named lth in model folder and put downloaded file CoNLL2009-ST-English-ALL.anna-3.3.srl-4.1.srl.model in it.<br>4. Create a folder named opennlp in model folder and put downloaded file en-sent.bin in it.<br>5. Compress the model folder to model.zip. |
| Virtual Machine. | 1. Start virtual machine.<br>2. Create a folder named koshik_test.<br>3. Copy enwiki-20160501-pages-articles.xml.bz2 to koshik_test folder.<br>4. Copy downloaded KOSHIK to koshik_test folder.<br>5. Copy model.zip to koshik_test folder. |
| Hadoop | 1. Under koshik_test folder, copy enwiki-20160501-pages-articles.xml.bz2 to Hadoop file system. |

## 4.3. Testing

The following show the steps to test KOSHIK to analyze the WIKI data. Steps 4, 5, and 6 are similar to those mentioned by Nugues (2014).

| Step | Description | Command |
|---|---|---|
| 1 | Import WIKI data into KOSHIK. | hadoop jar Koshik-1.0.1.jar se.lth.cs.koshik.util.Import -input /enwiki-20160501-pages-articles.xml -inputformat wikipedia -language eng -charset utf-8 -output /enwiki_avro |
| 2 | Start KOSHIK map-reduce jobs to analyze the data. | hadoop jar Koshik-1.0.1.jar se.lth.cs.koshik.util.EnglishPipeline -D mapred.reduce.tasks=12 -D mapred.child.java.opts=-Xmx8G -archives model.zip -input /enwiki_avro -output /enwiki_semantic |
| 3 | Import the analyzed result into Hive for querying. | CREATE EXTERNAL TABLE koshikdocs ROW FORMAT SERDE 'org.apache.hadoop.hive.serde2.avro.AvroSerDe' STORED AS INPUTFORMAT 'org.apache.hadoop.hive.ql.io.avro.AvroContainerInputFormat' OUTPUTFORMAT 'org.apache.hadoop.hive.ql.io.avro.AvroContainerOutputFormat' LOCATION '/hivetablekoshik' TBLPROPERTIES('avro.schema.url'='hdfs:///AvroDocument.avsc'); LOAD DATA INPATH '/enwiki_semantic/*.avro' INTO TABLE koshikdocs; |
| 4 | Query number of analyzed articles. | SELECT count(identifier) from koshikdocs; SELECT count(key) FROM (SELECT explode(ann) AS (key,value) FROM (SELECT ann FROM koshikdocs LATERAL VIEW explode(annotations.features) annTable as ann) annmap) decmap WHERE key='POSTAG' AND value LIKE 'NN%'; |
| 5 | Query number of sentences | SELECT count(ann) FROM koshikdocs LATERAL VIEW explode(annotations.layer) annTable as ann WHERE ann LIKE '%Sentence'; |
| 6 | Query number of nouns. | SELECT count(key) FROM (SELECT explode(ann) AS (key,value) FROM (SELECT ann FROM koshikdocs LATERAL VIEW explode(annotations.features) annTable as ann) annmap) decmap WHERE key='POSTAG' AND value LIKE 'NN%'; |

## 5. Findings: Advantages and Disadvantages of KOSHIK

KOSHIK provides an architecture which utilizes Hadoop and related tools as well as individual NLP tools and language models, and simplifies the construction of a whole NLP system. By adopting this architecture, people who work with an NLP system could focus on their own professional areas. For example, linguists could focus on creating effective language models to improve the accuracy of NLP, while algorithm designers could provide high performance NLP analysis components. When KOSHIK uses the sentence detection component from Mate and Stanford, it has the potential to process other human languages in the future, if the NLP tools continue to add new language models. Another advantage is that KOSHIK supports different kinds of document types, and it is given related APIs to expand the compatible document types. Moreover, with HDFS, KOSHIK has the ability to analyze a large sets of data with high input and output performance. The data processing can be passed on to MapReduce, which will provide more computing power as required.

Currently, KOSHIK has not yet become a mature product that is ready to be used by businesses.

First, it lacks support and documentation, and the learning curve for this tool is high. For example, this tool utilizes the language models of Mate, OpenNLP and Stanford CoreNLP. However, there are too many components for the user to discern which one KOSHIK is using for a particular task. Perhaps only its developers can see it from the source code. Secondly, since the last recorded change of code, this tool has not been maintained regularly, and there has not been any update for almost two years. It is uncertain whether this tool will have more functions and a stable version.

## 6. Conclusions

Natural language processing plays an important role in search engines, speech to text conversion tools, intelligent assistants, and artificial intelligence. It will continue to influence the user experience on the internet. With more and more data generated, there will be different kinds of data processed on Big Data platforms. Hadoop provides useful tools and has a mature ecosystem which is ideal for natural language processing. There are already some research reports, and software tools for natural language processing utilizing Hadoop. KOSHIK is one that provides an NLP architecture which utilizes Hadoop and Hive to process large amount of data. It is friendly for developers and linguists because it can separate these two types of work and allow each of them to focus on their own areas, thereby increasing the performance of NLP system. The architecture of KOSHIK is expandable; so it provides APIs to add new functions for the system. By utilizing OpenNLP, Mate and Stanford CoreNLP, KOSHIK supports different types of natural language processing. Documentation for KOSHIK is required to learn this tool, and to enable other developers to contribute toward it. However, because the source code has not been updated for a while, it is not ready for business use, but rather for personal testing and development.

## 7. Recommendations

Based on the architecture of KOSHIK and the process speed of large data, this architecture could increase processing speed by adopting Spark and GPU processing.

Spark is a cluster computing system maintained by Apache and it supports in-memory computing. This can improve the data analysis speed as high as possible, compared to the original MapReduce method in Hadoop (Zaharia, Chowdhury, Franklin, Shenker & Stoica, 2010). Based on the test results of Zaharia et al. (2010) where they used Spark to analyze a 39 GB wiki data, the query time using Spark was 0.5 to 1 second. The Hadoop query took 35 seconds; therefore, Spark is much faster than Hadoop.

At present, GPUs are successfully integrated into Hadoop and MapReduce frameworks which could increase the data processing speed (Yadav, Bhadoria & Suri, 2015). In addition, Yadav, Bhadoria and Suri (2015) found that there are some libraries such as JCUDA and Java Aparapi that provide APIs to interact with GPUs and extract better performance from them to support high performance computing within Hadoop.